\documentclass[10pt,twocolumn,letterpaper]{article}
\usepackage[pagenumbers]{cvpr}

\usepackage{graphicx}
\usepackage{amsmath}
\usepackage{amssymb}
\usepackage{booktabs}
\usepackage{comment}
\usepackage{float}
\usepackage{adjustbox}
\usepackage{algorithm}
\usepackage{algpseudocode}
\usepackage{multirow}
\usepackage{mathalpha}
\usepackage{hhline}

\usepackage[pagebackref,breaklinks,colorlinks]{hyperref}
\usepackage[accsupp]{axessibility} 

\usepackage[capitalize]{cleveref}
\crefname{section}{Sec.}{Secs.}
\Crefname{section}{Section}{Sections}
\Crefname{table}{Table}{Tables}
\crefname{table}{Tab.}{Tabs.}

\def\cvprPaperID{44}
\def\confName{CVPR}
\def\confYear{2023}

\begin{document}

\title{Identity-driven Three-Player Generative Adversarial Network for Synthetic-based Face Recognition}

\author{Jan Niklas Kolf$^{1,2}$, Tim Rieber$^{1}$, Jurek Elliesen$^{1,2}$, Fadi Boutros$^{1}$, Arjan Kuijper$^{1,2}$, Naser Damer$^{1,2}$\\
$^{1}$Fraunhofer Institute for Computer Graphics Research IGD, Darmstadt, Germany\\
$^{2}$Department of Computer Science, TU Darmstadt,
Darmstadt, Germany\\
Email: jan.niklas.kolf@igd.fraunhofer.de
}
\maketitle

\begin{abstract}
Many of the commonly used datasets for face recognition development are collected from the internet without proper user consent.
Due to the increasing focus on privacy in the social and legal frameworks, the use and distribution of these datasets are being restricted and strongly questioned. 
These databases, which have a realistically high variability of data per identity, have enabled the success of face recognition models. 
To build on this success and to align with privacy concerns, synthetic databases, consisting purely of synthetic persons, are increasingly being created and used in the development of face recognition solutions. 
In this work, we present a three-player generative adversarial network (GAN) framework, namely IDnet, that enables the integration of identity information into the generation process. 
The third player in our IDnet aims at forcing the generator to learn to generate identity-separable face images. 
We empirically proved that our IDnet synthetic images are of higher identity discrimination in comparison to the conventional two-player GAN, while maintaining a realistic intra-identity variation. 
We further studied the identity link between the authentic identities used to train the generator and the generated synthetic identities, showing very low similarities between these identities.
We demonstrated the applicability of our IDnet data in training face recognition models by evaluating these models on a wide set of face recognition benchmarks.
In comparison to the state-of-the-art works in synthetic-based face recognition, our solution achieved comparable results to a recent rendering-based approach and outperformed all existing GAN-based approaches. The training code and the synthetic face image dataset are publicly available \footnote{ \url{https://github.com/fdbtrs/Synthetic-Face-Recognition}}.

\end{abstract}

\vspace{-6mm}
\section{Introduction}
\label{sec:intro}
\vspace{-1mm}
Biometrics is a fast growing technology that recognizes people based on their physical or behavioral characteristics \cite{jain2004introduction}. One of the most commonly used modalities is the face, which is widely accepted by the population and can be captured without major hurdles \cite{spolaor2016biometric}. 
Face Recognition (FR) has become part of the everyday life of many end users, e.g. for unlocking smartphones. 
The breakthrough of robust FR systems in recent years is partly due to Deep Neural Networks\cite{pocketnet, ResNet}, which have, in combination with specific loss functions \cite{boutrostripletloss, elastic, ArcFace, CosFace}, significantly increased the recognition rates of these systems\cite{DBLP:journals/corr/abs-2102-00813, wangdeepface}.
However, DNN training requires a large amount and variation of data. 
Many of these datasets were compiled by crawling images from the Internet, thus raising both ethical and legal concerns \cite{DBLP:journals/ais/SmithM22, vgg2, guo2016ms}.

The issues of privacy and data autonomy are addressed in both an ethical and a legal framework. 
In the ethical context, Smith and Miller \cite{DBLP:journals/ais/SmithM22} relate privacy to the moral value of autonomy. This autonomy refers to the right to decide who is allowed to access personal information or data and how it is processed, including biometric data such as face images.
This standpoint is covered by the European Union's General Data Protection Regulation (GDPR) \cite{eu-2016/679}.

In the GDPR, biometric data is classified as special data requiring protection, thus it is subject to strong data protection regulations concerning acquirement and storage \cite{art9gdpr}. 
The processing of such data is restricted and the possibility of data withdrawal and control should be guaranteed  \cite{art7gdpr}.
However, these legal restrictions apply only to natural persons but not to synthetic identities such as characters drawn by artists or images of subjects generated by a machine \cite{art4gdpr}.

To address ethical and legal concerns, the field of synthetic data for biometric development is advancing increasingly \cite{DBLP:journals/ieeesp/Emam20, lopez_elbi_2022}. 
The recent studies in this direction aimed at creating datasets for the development of FR \cite{usynthface,boutros2022sface,bae2023digiface}, along with other FR system components, without the need for images of authentic identities \cite{DBLP:conf/cvpr/DamerLFSPB22,DBLP:journals/corr/abs-2303-02660,DBLP:conf/icpr/BoutrosDK22,DBLP:conf/icb/HuberBLRRDNGSCT22}.
The generation of synthetic data is primarily driven by Generative Adversarial Networks (GAN). These are able to generate photo-realistic results based on  training a two-player minimax game between a generator and a discriminator. Various approaches using GAN for the generation of synthetic identities with respect to FR have been presented very recently, such as DigiFace-1M \cite{bae2023digiface}, SynFace \cite{synface}, SFace \cite{boutros2022sface} or USynthFace \cite{usynthface}. 
Although these approaches presented very promising FR results, they have some limitations. 
SFace \cite{boutros2022sface} suffers from relatively low identity separability in comparison to authentic data, i.e. the identities are not highly distinct and thus genuine (same identity) comparisons might be in some cases confused to be imposter (different identity) comparisons. 
SynFace \cite{synface} mixed up authentic and synthetic data in FR training, aiming at increasing intra-class variations. USynthFace proposed to mitigate the low intra-class variations by proposing aggressive data augmentation  to train unsupervised FR models. However, USynthFace is limited to unsupervised learning. DigiFace-1M utilized a digital rendering pipeline to construct synthetic images.  DigiFace-1M \cite{bae2023digiface} is extremely computationally costly, limiting its ability on generating large-scale synthetic data and such high computational demand might not be available for research, along with the low FR performance it produces without the dependency on sophisticated augmentation.

In this work, we propose IDnet, which extends the traditional class conditioned GAN by adding an additional identity-dedicated third player (ID-3) to the minimax game. This component overcomes the drawbacks of the class conditioned GAN (as represented in SFace \cite{boutros2022sface}) by enforcing specific identity-discriminant information in the generation process. 
In several experiments, we show that this additional component causes the generator of the GAN to produce images of synthetic identities that stronger resemble the distribution of authentic data, both in terms of class separability and intra-class variance in comparison to the considered baseline SFace.  We show that FR systems trained on synthetic datasets generated with IDnet outperform those trained on datasets synthesized with our baseline, SFace, and other GAN-based synthetic datasets.
\vspace{-2mm}
\section{Related Work}
\label{sec:related_work}
\vspace{-1mm}
Current state-of-the-art (SOTA) FR models such as ElasticFace \cite{elastic}, ArcFace \cite{ArcFace}, and CosFace \cite{CosFace} are trained on authentic datasets such as CASIA-Webface \cite{yi2014learning} and MS1MV2 \cite{guo2016ms,ArcFace}. Such models regularly achieve new record breaking results on the authentic benchmarks \cite{elastic} such as LFW \cite{lfw} or AgeDB-30 \cite{agedb}. With the goal of generating identity-specific face images, various methods are proposed in the literature, with most using GAN as the basis for generating synthetic data. 
Marriott \etal \cite{identitygan} were among the first to evaluate the capabilities of GAN for identity-based applications such as facial recognition. In their work, they have shown that GAN can be used to create synthetic identities that are not included in the training set. By introducing a special triplet loss, the authors were also able to increase the identity disentanglement, which means that the identity information can be separated from the other image properties in the generation process. DiscoFaceGAN \cite{discofacegan} explored an approach to create facial images from non-existing people with the use of multiple disentangled features as input for the generator. These features include identity, pose, illumination, and expression. To ensure that these features are disentangled, a contrastive loss is used. 

Towards training FR based on synthetic data, SynFace \cite{synface} aimed at training FR models with the use of synthetic data. It investigated the performance of DiscoFaceGAN in terms of intra-class variance and the domain gap between real and generated images. SynFace improved these aspects of DiscoFaceGAN by introducing identity and domain mixups. Mixup uses combinations of image features to generate new faces.
FaceID-GAN \cite{faceidgan} introduced a classification network as an additional constraint to the GAN training.
FaceID-GAN aims at generating variations of input authentic images rather than generating images of synthetic identities.
SFace \cite{boutros2022sface} proposed class-conditional synthetic GAN for class-labeled synthetic image generation. The authors utilize synthetic data to train supervised FR models, achieving  very promising verification accuracies.
USynthFace \cite{usynthface} showed that by using unlabeled synthetic data, a FR model can be trained successfully using contrastive learning and aggressive data augmentation.
By rendering $3$d human models that are varied in appearance, shape, and accessories, among others, the authors of DigiFace-1M \cite{bae2023digiface} have created a synthetic dataset that exhibits high variability in the data generated for a synthetic identity. This allowed them to train FR models based on their proposed data, achieving relatively high verification accuracies when combined with sophisticated rendering technique. 
However, such a digital image rendering approach is computationally costly, limiting their ability in generating large-scale synthetic datasets. 

While there has been great progress in the area of synthetic FR datasets generated by GAN, previous synthetic data used for FR training either suffer from low intra-class diversity or are of low identity separability.
In this work, we propose a novel approach for generating synthetic images that are identity separable while containing realistic intra-class variations. 
\begin{figure*}[ht]
\centering
\includegraphics[width=0.83\textwidth]{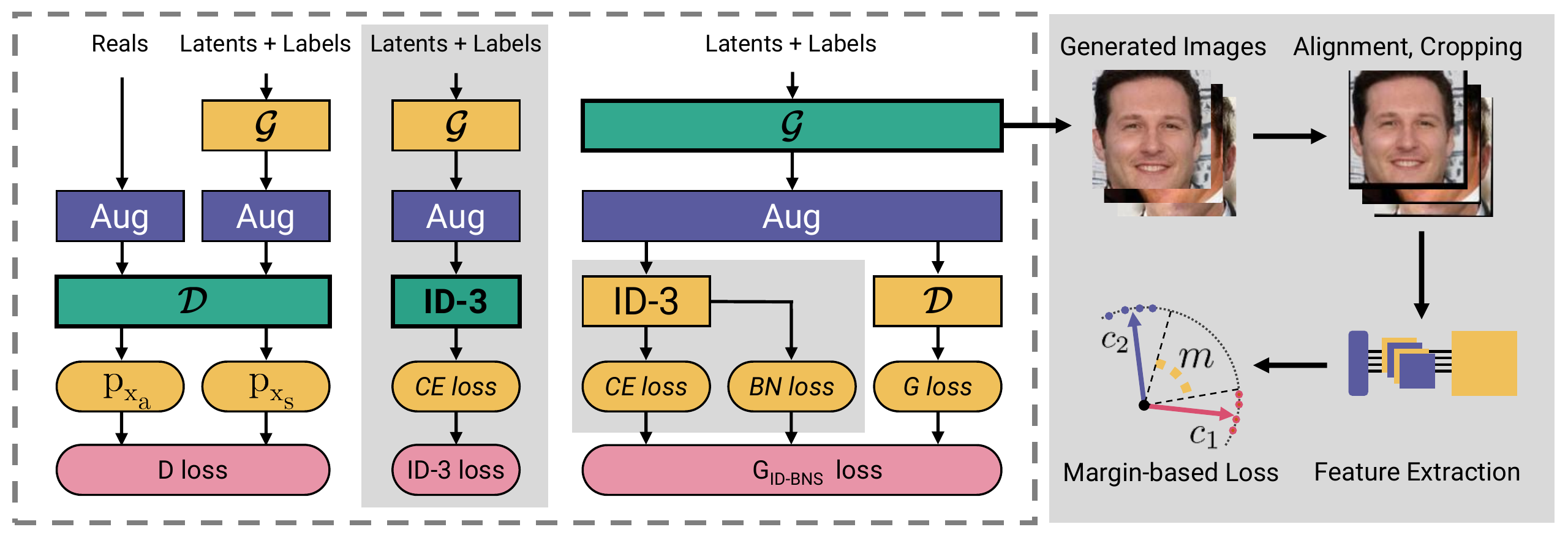}
\vspace{-3mm}
\caption{An overview of IDnet with the two-player StyleGAN2-ADA architecture as well as the third-player ID-3 extension presented in this paper, which are highlighted in gray. Three models are trained at the same time, whereby the individual components, which are updated by the appropriate loss, are marked as black bordered green boxes. The $D$ loss is defined in \cref{eq:loss_discriminator}, $G$ loss in \cref{eq:loss_g_solo}, ID-3 loss in \cref{eq:id3_loss},  and $\text{G}_{ID-BNS}$ loss in \cref{eq:loss_g}.
Using the trained generator, a synthetic face dataset is created, which is aligned and cropped, and on which a FR model with margin-penalty based loss function is trained.
}
\label{fig:pipeline}
\vspace{-5mm}
\end{figure*}
\vspace{-2mm}
\section{Methodology}
\label{sec:methodology}
\vspace{-1mm}
We propose in this work an identity-conditioned generative model based on a Generative Adversarial Network (GAN) \cite{gan}, namely IDnet. 
The conventional GAN are based on optimizing a minimax game between two players, the generator and the discriminator. The generator aims at fooling the discriminator by generating synthetic images that are similar to authentic images. The discriminator tries to distinguish between synthetic and authentic images \cite{gan, stylegan2}. 
As these two models compete against each other, the discriminator pushes the generator to learn to estimate the probability distribution of authentic training data, enabling the generation of realistic synthetic samples. 
We propose in this work to extend this game by a third player, an identity model, ID-3. The  ID-3 aims at pushing the generator not only to learn the underlying probability distribution of authentic data but also to learn the identity information encoded in training samples, and thus, generating identity-separable samples of synthetic identity.
\vspace{-1mm}
\subsection{Two-Player GAN}
\vspace{-1mm}
We start here by introducing the needed formulation and background to our proposed solution. GAN was originally proposed by Goodfellow \etal \cite{gan}, aiming at generating synthetic data by sampling latent vectors from a given random distribution e.g. Gaussian distribution, and feeding it into the generator.
GAN training is based on training two players, the generator $\mathcal{G}$ and the discriminator $\mathcal{D}$ \cite{gan, stylegan2}.
$\mathcal{G}$ receives a latent vector $\mathbf{z}$ as input and outputs a synthetic image $\mathbf{x}_{s} = \mathcal{G}(\mathbf{z})$. The training objective of $\mathcal{G}$ is to learn to generate synthetic images of the same probability distribution of authentic training data.
The training objective of $\mathcal{D}$ is to distinguish between authentic images $\mathbf{x}_a$ and $\mathbf{x}_{s}$.
In a binary classification problem, $\mathcal{D}$ trains to estimate the probability of input $\mathbf{x}$ being from authentic or synthetic distributions. Formally, $\mathcal{D}$'s training objective can be defined as follows:
\vspace{-1mm}
\begin{equation}
    L_{D}=log(1+e^{-p_{x_a}}) + log(1+e^{p_{x_{s}}}),
    \label{eq:loss_discriminator}
\end{equation}
where  $p_{x_{a}}=\mathcal{D}(\mathbf{x}_a)$ and $p_{x_{s}}=\mathcal{D}(\mathbf{x}_s)$ are the probabilities of $\mathbf{x}_a$ and $\mathbf{x}_s$ being from  authentic or synthetic distributions, respectively.

$\mathcal{G}$ aims at fooling $\mathcal{D}$ by generating realistic samples that would minimize the logarithm of the inverse probability $p_{x_{s}}$ (predicted by $\mathcal{D}$). 
As $\mathcal{D}$ and $\mathcal{G}$ compete against each other, the training loss of $\mathcal{G}$ is defined as follows:
\vspace{-1mm}
\begin{equation}
    L_{G}=log(1+e^{-p_{x_{s}}}),
    \label{eq:loss_g_solo}
\end{equation}
where $p_{x_{s}}$ is $\mathcal{D}$ prediction i.e. prediction of $\mathbf{x}_s$ is being from synthetic distribution.
The only condition on conventional GAN is that the generated images tend to have a similar probability distribution to the authentic training data.
Several previous works \cite{discofacegan, stylegan, stylegan2} proposed to extend conventional GAN with conditional mechanisms, enabling synthetic image generation with certain attributes. 
One of the widely used conditional mechanisms is class-conditional GAN. In this case, class labels are used as additional input to $\mathcal{G}$ and $\mathcal{D}$ to force $\mathcal{G}$ to learn to generate synthetic images of a specific class.  SFace \cite{boutros2022sface}, which is based on class-conditional StyleGAN2-ADA \cite{stylegan2-ada}, trained GAN under class-conditional settings to generate face images of a specific class label. In this case, a class label ($c$) is embedded into a 512-D vector and then is concatenated with the input latent vector $\mathbf{z}$ ($\mathbf{z}$ is of 512-D) to generate class-related synthetic images ($\mathbf{x}_{s}=\mathcal{G}(\mathbf{z},c)$).  In  class-conditional $\mathcal{D}$,  a class label ($c$) is embedded into a 512-D vector and then concatenated with the final embedding layer of $\mathcal{D}$.
The reported results by SFace demonstrated that SFace synthetic images are realistic with large intra-class variations. 
However, as we present in this paper (\Cref{sec:results}), the generated images by SFace suffer from relatively low identity separability which might lead to less optimal face verification accuracies when such synthetic data is used to train FR.
In such a training paradigm, GAN is guided to learn to generate synthetic images from a specific class, however, not from a specific identity. To overcome this challenge, we introduce a third player to GAN, an identity network ID-3. Our new player aims at guiding $\mathcal{G}$ to learn to generate synthetic images that are highly identity-separable.
\vspace{-1mm}
\subsection{Identity Network as a Third-Player}
\vspace{-1mm}
The third player in the minimax game is noted as ID-3 which acts as a FR model.  ID-3 is trained with a margin-penalty softmax loss that incorporates margin penalty in softmax loss to push training samples to be close to their class centers and far away from other class centers. Specifically, we use CosFace \cite{CosFace} loss to train ID-3. Also, ID-3 loss is used as an additional loss of $\mathcal{G}$ loss to guide $\mathcal{G}$ to generate synthetic images that are identity separable. The ID-3 loss on a batch of synthetic images $\mathbf{x}_s$ are given as follows: 
\begin{equation}
\resizebox{0.9\linewidth}{!}{$
    L_{ID-3}=-\frac{1}{N}  \sum\limits_{i \in N}log \frac{e^{s (cos(\theta_{y_i})-m)}}{ e^{s(cos(\theta_{y_i})-m)} +\sum\limits_{j=1 , j \ne y_i}^{C}  e^{s\ cos(\theta_{j})}},
$}
\label{eq:id3_loss}
\end{equation}
where $m$ is a margin penalty, $s$ is a scaling term, $N$ is the batch size, $C$ is the number of classes and $\theta_{y_i}$ is the angle between sample $y_i$ and its $i-th$ class center. 
Synthetic samples in the early stage of GAN training tend to be less realistic with low identity information encoded in the generated images. 
At an early stage of IDnet training, pushing such images to their class centers with a large margin penalty value will affect ID-3 training stability.
Thus, we propose a progressive margin-penalty value to stabilize ID-3 training. Initially, $m$ is set to $m=0$ and it is incrementally increased by a small value of $0.05$ every 7 epochs with a maximum of $0.35$ \cite{CosFace}.
During our three-player game, only synthetic images $\mathbf{x}_s$ (generated by $\mathcal{G}$) are used to train ID-3, and the ground-truth labels are their corresponding class labels ($c$). We used a pretrained ResNet-50 on CASIA-WebFace \cite{yi2014learning} with CosFace \cite{CosFace} as a backbone for our ID-3. We froze all the network weights and only train the weights of the classification layer.
The loss function of $\mathcal{G}$ in our three-player game is given by:
\begin{equation}
     L_{G_{ID}}=L_{G} + L_{ID-3}.
     \label{eq:L-G-ID}
\end{equation}

\paragraph{Domain Adaptation (DA)}
We propose in this work to further minimize the domain gap between synthetic and authentic data distributions by matching Batch Normalization Statistics ($BNS$) of authentic and synthetic data. Following \cite{xu2020generative}, we first extracted means $\mu_{a}$ and standard deviations $\sigma_{a}$ of all batch normalization ($BN$) layers of the ID-3 backbone (trained on authentic data). During our three-player GAN training, we calculated the means $\mu_{s}$ and the standard deviations $\sigma_{s}$ of all $BN$ layers by passing a batch of synthetic data into ID-3 and then extracted $BN$ statistics. Finally, we calculate the difference between $BN$ statistics from authentic and synthetic data as follows: 
\begin{equation}
    L_{BNS}(\mu_s, \sigma_s) = \sum_{l \in \text{BNL}}\left \| \mu_s^l-\mu_a^l \right \|_2^2+\left \| \sigma_{s}^l-\sigma_{a}^l \right \|_2^2,
\end{equation}
where $\text{BNL}$ are BN layers of ID-3.
The $L_{BNS}$ is used as an additional loss to $G$ loss (defined in \cref{eq:L-G-ID}). The final loss $L_{G_{ID-BNS}}$ for $\mathcal{G}$ in our three-player GAN is given by:
\begin{equation}
    L_{G_{ID-BNS}} = L_{G_{ID}} + \lambda*L_{ID-3} + \kappa*L_{BNS},
\label{eq:loss_g}
\end{equation}
where $\lambda$ and $\kappa$ are weighting terms for $L_{ID-3}$ and $L_{BNS}$, respectively.
The training algorithm of our IDnet is shown in \Cref{algo:training}, the pipeline is shown in \Cref{fig:pipeline}. 

\begin{algorithm}
\begin{algorithmic}
\caption{Three-Player GAN IDnet Training Loop}
\label{algo:training}
\small
\State $m \gets 0$
\State $m_{\delta} \gets 0.05$
\For{$epoch$ in $epochs$}
    \State $m \gets m+m_{\delta} \text{ if } epoch \% 7 = 0$
    \For{$batch$ in $trainingset$}
        \State $\mathbf{z} \gets \text{Sampled from Gaussian Dist.}$
        \State $c \gets \text{Randomly sampled from C}$
        \State $\mathbf{x}_{s} \gets \mathcal{G}(\mathbf{z}, c)$
        \State $p_{x_{s}} \gets \mathcal{D}(\mathbf{x}_{s}, c)$
        \State $\mu_s, \sigma_s \gets BNS(\text{ID-3}(\mathbf{x}_{s}))$
        \State $L_{G} \gets L_{G}(p_{x_{s}})$
        \State $L_{ID-3} \gets L_{ID-3}(p_c, c)$
        \State $L_{BNS} \gets L_{BNS}(\mu_s,\sigma_s)$
        \State $backward(\mathcal{G},  L_{G_{ID-BNS}}(L_{G}, L_{ID-3}, L_{BNS}))$
        \State $\mathbf{x}_{a}, c \gets \text{batch}$
        \State $p_{x_{a}} \gets \mathcal{D}(\mathbf{x}_{a}, c)$
        \State $p_{x_{s}} \gets \mathcal{D}(\mathbf{x}_{s}, c)$
        \State $backward(\mathcal{D}, L_{D}(p_{x_{a}}, p_{x_{s}}))$
        \State $p_{c} \gets \text{ID-3}(\mathbf{x}_{s}), c)$
        \State $backward(\text{ID-3}, L_{ID-3}(p_{c},c))$
        \State $update(\mathcal{D})$
        \State $update(\mathcal{G})$
        \State $update(\text{ID-3})$
        \State $zerograd()$
    \EndFor
\EndFor
\end{algorithmic}
\end{algorithm}

\vspace{-4mm}
\section{Experimental Setup}
\label{sec:experimental_setup}
\vspace{-1mm}
\paragraph{Dataset:}
We use CASIA-Webface \cite{yi2014learning} to train our three-player GAN (as described in \Cref{sec:methodology}).
CASIA-Webface consists of $494,414$ authentic images from $10,572$ different identities \cite{yi2014learning}. 
Multi-Task Cascaded Convolutional Networks (MTCNN) \cite{mtcnn} is used to extract the facial landmarks based on which the faces are aligned and cropped to the training size of $128 \times 128 \times 3$, as described in \cite{ArcFace}.
\vspace{-4mm}
\paragraph{IDnet Training Settings:}
In this work we utilize StyleGAN2-ADA \cite{stylegan2-ada}, as used in SFace \cite{boutros2022sface}, as our two-player GAN foundation.
StyleGAN2-ADA improves StyleGAN2 \cite{stylegan2} by adaptive dataset augmentations (ADA). These augmentation methods are applied to the authentic and synthetic images, allowing more stable training, especially when a small dataset is used for training. StyleGAN2-ADA is extended with the additional identity network and respective losses of our three-player minimax game.\\
The generator and discriminator of StyleGAN2-ADA are based on Progressive GANs \cite{ProgGAN} as originally presented in \cite{stylegan2-ada}. The dimensionality of the style vectors and noise is set to $512$. The dimensionality of the conditioning and therefore the number of synthetic identities is set to be equal to the number of identities in CASIA-Webface, with $C=10,572$, following \cite{boutros2022sface}. The mapping architecture consists of $8$ fully connected layers. The activation function for layers in the generator is Leaky ReLU \cite{relu} with $\alpha=0.2$. The generator outputs images of size $128\times128\times3$. We follow  the training setup described in \cite{boutros2022sface} and  \cite{stylegan2-ada}. The loss function of the discriminator is the non-saturating loss as described in \cite{gan} with $R_1$ regularization \cite{DBLP:conf/icml/MeschederGN18}. For the additional components of the generator's loss function (see formula \ref{eq:loss_g}), the weights $\lambda=0.05$ and $\kappa=0.1$ are chosen. As optimizer for the generator and discriminator, Adam is used with the parameters $\beta_1=0$, $\beta_2=0.99$ and $\epsilon=10^{-8}$. The learning rate is set to $0.0025$.
The backbone of ID-3 is ResNet-50 \cite{ResNet} with an embedding size of $512$.
The synthetic images are resized to  $112\times112\times3$ using bilinear interpolation to match ID-3 input size.
We set $s$ to $64$ \cite{CosFace} in $L_{ID-3}$ and use  a progressively growing margin as described in \Cref{sec:methodology}.
The three-player GAN is trained for 50 epochs.
The optimizer of ID-3 is Stochastic Gradient Descent with a learning rate of $0.02$ and momentum of $0.9$.
The minibatch size is set to $32$. IDnet is implemented using PyTorch based on the official PyTorch implementation \cite{sg2pytorch} of StyleGAN2-ADA. The models are trained on a Linux machine (Ubuntu 20.04.2 LTS) with Intel(R) Xeon(R) Gold 5218 CPU 2.30GHz, 512GB RAM and 4 Nvidia GeForce RTX 6000 GPUs.
\vspace{-4mm}
\paragraph{Class separability:}
In order to train a FR model with synthetic data, it is assumed that the synthetic data should resemble the distribution of authentic data. Authentic identities can be separated by FR systems \cite{elastic}, to a certain degree, as they each have characteristics that make them distinguishable from other authentic identities. 
Higher separability between genuine and imposter comparison scores indicates higher identity discrimination.
Thus, with higher identity discrimination the False Non-Match Rates (FNMR) and False Match Rate (FMR) are lower at a chosen decision threshold than those error rates that result from data with low identity discrimination. FNMR and FMR are based on the ISO/IEC 19795-1 \cite{mansfield2006information} standard. 
We use FMR100, FMR1000 and Equal Error Rate (EER) as class separability evaluation metrics.
The FMR100 gives the lowest FNMR at the operation point FMR $\leq 1.0\%$, the FMR1000 gives the lowest FNMR at FMR $\leq 0.1\%$. The Equal Error Rate (EER)  \cite{mansfield2006information} is FMR or FNMR at the operation point at which they are equal.

Based on that, using EER, FMR, and FNMR, we can measure how the identity discrimination in our IDnet data compares to the authentic data and to the SFace baseline.
For evaluation, $10$ images per synthetic identity are generated for both IDnet and SFace. This results in two synthetic datasets with $105,720$ images each.
For each synthetic image as well as each image from CASIA-Webface, a $512$ dimensional embedding is extracted using two pre-trained FR models, ResNet-100 trained with the Elastic-ArcFace loss \cite{elastic}, and the CurricularFace loss \cite{curricularface}. Individual pairwise comparisons are formed from the embeddings. The cosine similarity is used as a similarity measure. Comparisons of two face embeddings of the same identity (class label) are genuine comparisons, while comparisons between different identities are imposter comparisons.\\
Each image of an identity is compared to all other images of that respective identity as a genuine comparison. This image is also compared to $100$ randomly sampled images of other identities, resulting in $100$ imposter scores for each image. 
The genuine and imposter scores distributions are plotted for authentic CASIA-Webface, synthetic SFace and IDnet, respectively. For each method, the EER, FMR100, FMR1000 are calculated and reported.
\vspace{-4mm}
\paragraph{Intra-Class Variance:}
To match the distribution of authentic face data it is not only necessary to achieve a similar class separability, but also to resemble the intra-class variability of authentic identities. Among other things, variations in the appearance or different lighting conditions can change the visual appearance of an authentic identity. A FR model must learn how to determine the identity even in the presence of large visual variation. Therefore the generated synthetic data should also have realistic variation within a synthetic identity. 
To evaluate the intra-class variance of a dataset, the variance in the face embeddings of one identity is calculated:
\vspace{-2mm}
\begin{equation}
    \text{ICV} = \frac{1}{N*(N-1)} \sum_{i=1}^{N} \sum_{j=i+1}^{N} \frac{1}{D} \sum_{k=1}^{D} \left \| f_{i,k} - f_{j,k} \right \|,
\label{formula:ICV}
\end{equation}
where $N$ is the number of images of one identity and $D$ is the feature dimensionality of a face embedding. The ICV is calculated for every identity in the given dataset and summed up.
The ICV score is calculated for the authentic dataset CASIA-Webface and the synthetic datasets generated by SFace and IDnet.
\paragraph{The Link between Authentic and Synthetic Identities:}
To ensure the privacy motivation behind the generated synthetic data, there should be no major identity linkage between authentic identities from the IDnet training dataset, CASIA-Webface, and the generated synthetic identities from our proposed IDnet. To investigate the possible identity linkage, an additional experiment is conducted following those proposed in \cite{boutros2022sface}.

For each identity of CASIA-Webface, the first two images are selected as reference images. Probe images with the same class label are selected from CASIA-Webface, SFace, and IDnet. The reference images are compared to the probes of the respective dataset, resulting in score distributions with CASIA-Webface references vs. CASIA-Webface probes, CASIA-Webface references vs. SFace probes and CASIA-Webface references vs IDnet probes. The stronger the identity linkage between CASIA-Webface and SFace or IDnet, the stronger the distributions of SFace probes and IDnet probes shifts towards the CASIA-Webface genuine probes distribution.
The comparison scores, here and in the intra-class variance analyses, are a result from comparing face templates produced by a publicly released ResNet-100 trained on Elastic-ArcFace loss \cite{elastic}.
\vspace{-4mm}
\paragraph{Face recognition based on IDnet:}
The primary use of our synthetic data is to train a FR model with more shareable, scalable, and privacy-friendly data. Therefore, it is important to investigate to what extent a FR model trained on the synthetic data is able to perform on benchmarks consisting of authentic images. For this purpose, ResNet-50 models are trained with CosFace on synthetic data generated by our IDnet with parameters following \cite{boutros2022sface, elastic}. The chosen parameter for dropout is $0.4$, as embedding size $512$ is used. The CosFace margin is set as $m=0.35$ and the scale parameter $s$ to $s=64$. The minibatch size is $512$. Stochastic Gradient Descent with a learning rate of $1e-1$ is used as an optimizer. Momentum is $0.9$ and weight decay $5e-4$. The learning rate is divided by ten at the 22nd, 30th and 40th epochs, all models are trained for 40 epochs. 
Per identity, $10$, $20$, $40$, $50$ and $60$ images are randomly generated as proposed in \cite{boutros2022sface}. All images are aligned with the previously mentioned MTCNN algorithm \cite{mtcnn}.
During the training, the input images are augmented using the random augmentation methods introduced in \cite{DBLP:conf/fgr/BoutrosKFKD23}. As authentic image benchmarks Labeled Faces in the Wild (LFW) \cite{lfw}, Cross-age LFW (CALFW) \cite{calfw}, Cross-Pose LFW (CPLFW) \cite{cplfw}, Celebrities in Frontal-Profile in the Wild (CFP-FP) \cite{cfpfp} and AgeDB30 \cite{agedb} are used. We follow the evaluation protocol specified in the benchmarks. For all benchmarks, the verification performance is given as accuracy $[\%]$. We compare our work to recent works of SFace \cite{boutros2022sface}, USynthFace \cite{usynthface}, SynFace \cite{synface} and DigiFace-1M \cite{bae2023digiface} that have utilized ResNet-50 as a backbone model as well.
\begin{figure*}[h]
    \captionsetup[subfigure]{justification=centering}
     \centering
     \begin{subfigure}[t]{0.09\linewidth}
         \centering
         \includegraphics[width=\textwidth]{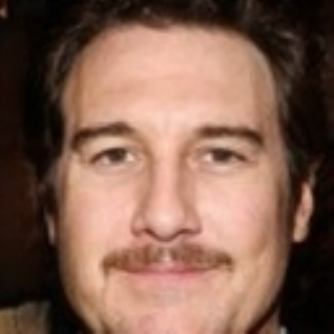}
         \caption{SFace Reference\\} 
         \label{fig:844_cond_ref}
     \end{subfigure}
     \begin{subfigure}[t]{0.09\linewidth}
         \centering
         \includegraphics[width=\textwidth]{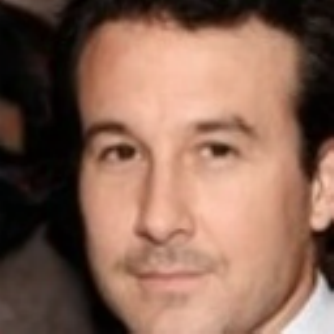}
         \caption{SFace Probe \#1 ($0.506$)}
         \label{fig:844_cond_ref1}
     \end{subfigure}
     \begin{subfigure}[t]{0.09\linewidth}
         \centering
         \includegraphics[width=\textwidth]{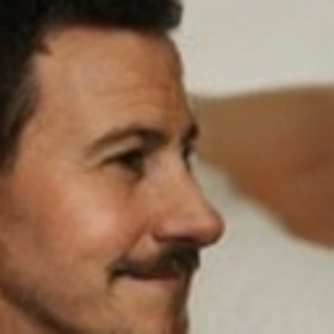}
         \caption{SFace Probe \#2 ($0.255$)}
         \label{fig:844_cond_ref2}
     \end{subfigure}
     \begin{subfigure}[t]{0.09\linewidth}
         \centering
         \includegraphics[width=\textwidth]{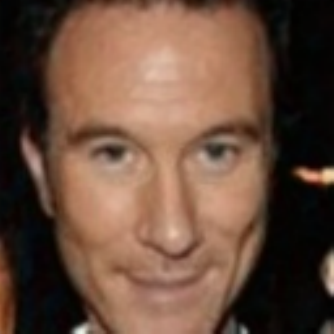}
         \caption{SFace Probe \#3 ($0.302$)}
         \label{fig:844_cond_ref3}
     \end{subfigure}
     \begin{subfigure}[t]{0.09\linewidth}
    \centering
    \includegraphics[width=\textwidth]{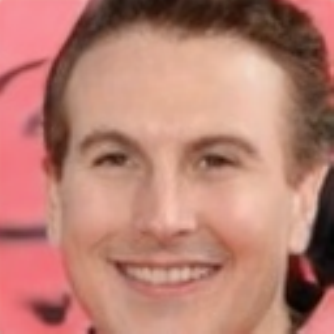}
    \caption{SFace Probe \#4 ($0.214$)}
    \label{fig:844_cond_ref4}
    \end{subfigure}
    \begin{subfigure}[t]{0.09\linewidth}
    	\centering
    	\includegraphics[width=\textwidth]{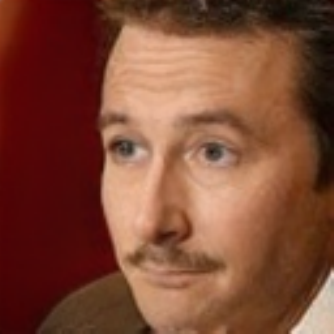}
    	\caption{SFace Probe \#5 ($0.285$)}
    \label{fig:844_cond_ref5}
    \end{subfigure}
    \begin{subfigure}[t]{0.09\linewidth}
    	\centering
    	\includegraphics[width=\textwidth]{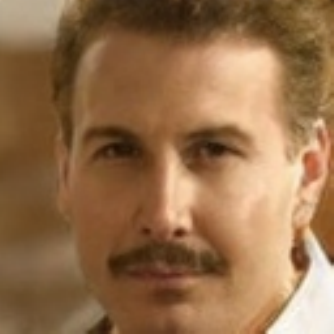}
    	\caption{SFace Probe \#6 ($0.403$)}
    \label{fig:844_cond_ref6}
    \end{subfigure}
    \begin{subfigure}[t]{0.09\linewidth}
    	\centering
    	\includegraphics[width=\textwidth]{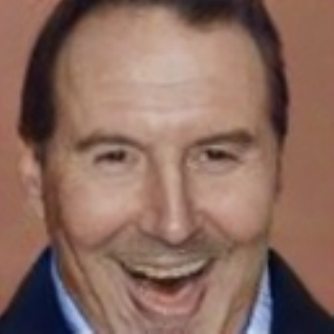}
    	\caption{SFace Probe \#7 ($0.266$)}
    \label{fig:844_cond_ref7}
    \end{subfigure}
    \begin{subfigure}[t]{0.09\linewidth}
    	\centering
    	\includegraphics[width=\textwidth]{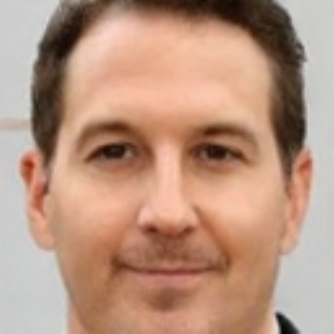}
    	\caption{SFace Probe \#8 ($0.427$)}
    \label{fig:844_cond_ref8}
    \end{subfigure}
    \begin{subfigure}[t]{0.09\linewidth}
    	\centering
    	\includegraphics[width=\textwidth]{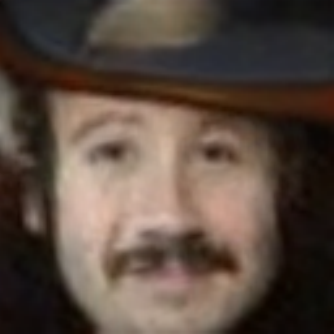}
    	\caption{SFace Probe \#9 ($0.190$)}
    \label{fig:844_cond_ref9}
    \end{subfigure}
    
    %
    %

    \begin{subfigure}[t]{0.09\linewidth}
         \centering
         \includegraphics[width=\textwidth]{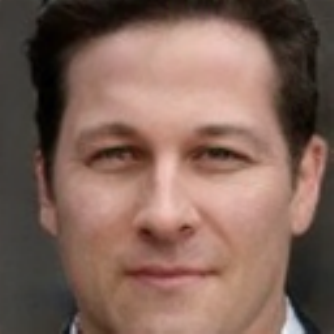}
         \caption{IDnet Reference}
         \label{fig:844_idcond_ref}
    \end{subfigure}
    \begin{subfigure}[t]{0.09\linewidth}
         \centering
         \includegraphics[width=\textwidth]{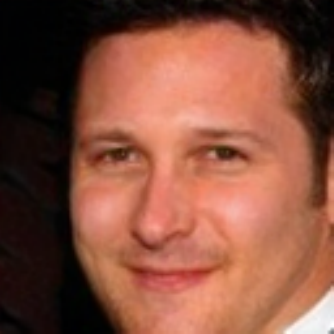}
         \caption{IDnet \\Probe \#1 ($0.458$)}
         \label{fig:844_idcond_ref1}
    \end{subfigure}
    \begin{subfigure}[t]{0.09\linewidth}
         \centering
         \includegraphics[width=\textwidth]{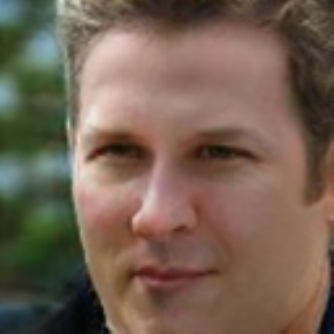}
         \caption{IDnet  Probe \#2 ($0.501$)}
         \label{fig:844_idcond_ref2}
    \end{subfigure}
    \begin{subfigure}[t]{0.09\linewidth}
         \centering
         \includegraphics[width=\textwidth]{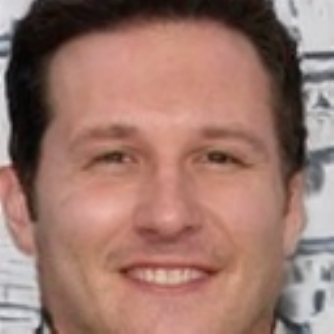}
         \caption{IDnet  Probe \#3 ($0.428$)}
         \label{fig:844_idcond_ref3}
    \end{subfigure}
    \begin{subfigure}[t]{0.09\linewidth}
        \centering
        \includegraphics[width=\textwidth]{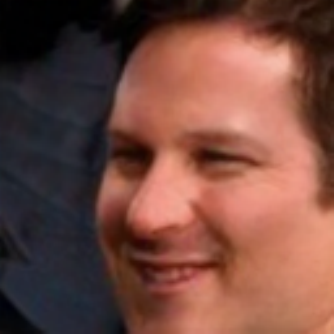}
        \caption{IDnet Probe \#4 ($0.346$)}
        \label{fig:844_idcond_ref4}
    \end{subfigure}
    \begin{subfigure}[t]{0.09\linewidth}
    	\centering
    	\includegraphics[width=\textwidth]{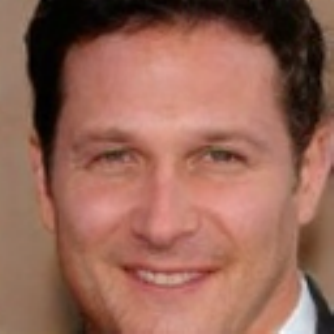}
    	\caption{IDnet Probe \#5 ($0.474$)}
    \label{fig:844_idcond_ref5}
    \end{subfigure}
    \begin{subfigure}[t]{0.09\linewidth}
    	\centering
    	\includegraphics[width=\textwidth]{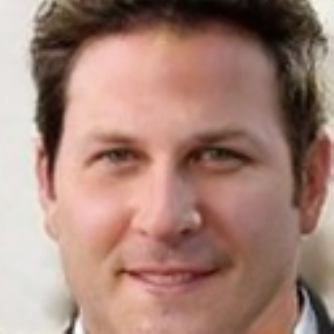}
    	\caption{IDnet Probe \#6 ($0.514$)}
    \label{fig:844_idcond_ref6}
    \end{subfigure}
    \begin{subfigure}[t]{0.09\linewidth}
    	\centering
    	\includegraphics[width=\textwidth]{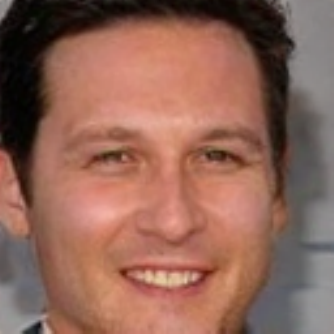}
    	\caption{IDnet Probe \#7 ($0.432$)}
    \label{fig:844_idcond_ref7}
    \end{subfigure}
    \begin{subfigure}[t]{0.09\linewidth}
    	\centering
    	\includegraphics[width=\textwidth]{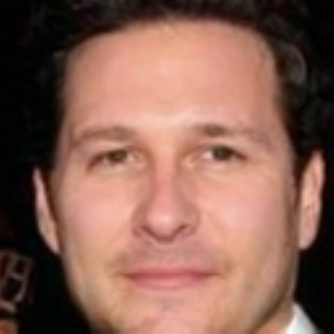}
    	\caption{IDnet Probe \#8 ($0.547$)}
    \label{fig:844_idcond_ref8}
    \end{subfigure}
    \begin{subfigure}[t]{0.09\linewidth}
    	\centering
    	\includegraphics[width=\textwidth]{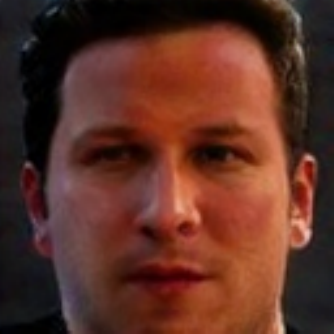}
    	\caption{IDnet Probe \#9 ($0.500$)}
    \label{fig:844_idcond_ref9}
    \end{subfigure}     
    \vspace{-3mm} 
      \caption{Example images of a specific identity (\#844), generated by SFace (a-j) and our IDnet method (k-t). Image (a) is compared to probe images (b-j), likewise image (k) is compared pairwise to (l-t). The cosine similarity scores are given for each comparison. The average score for SFace is $0.316$ and for IDnet it is $0.467$, respectively. From the similarity scores it is evident, that IDnet generates images with significantly better distinct identity information than SFace.}
    \label{fig:id844_comparison}
    \vspace{-3mm}
    
\end{figure*}

\vspace{-2mm}
\section{Results}
\label{sec:results}
\vspace{-2mm}
\paragraph{Investigating Class Separability:}

\begin{figure*}[h!]
    \captionsetup[subfigure]{justification=centering}
     \centering
     \begin{subfigure}[t]{0.3\linewidth}
         \centering
         \includegraphics[width=\textwidth]{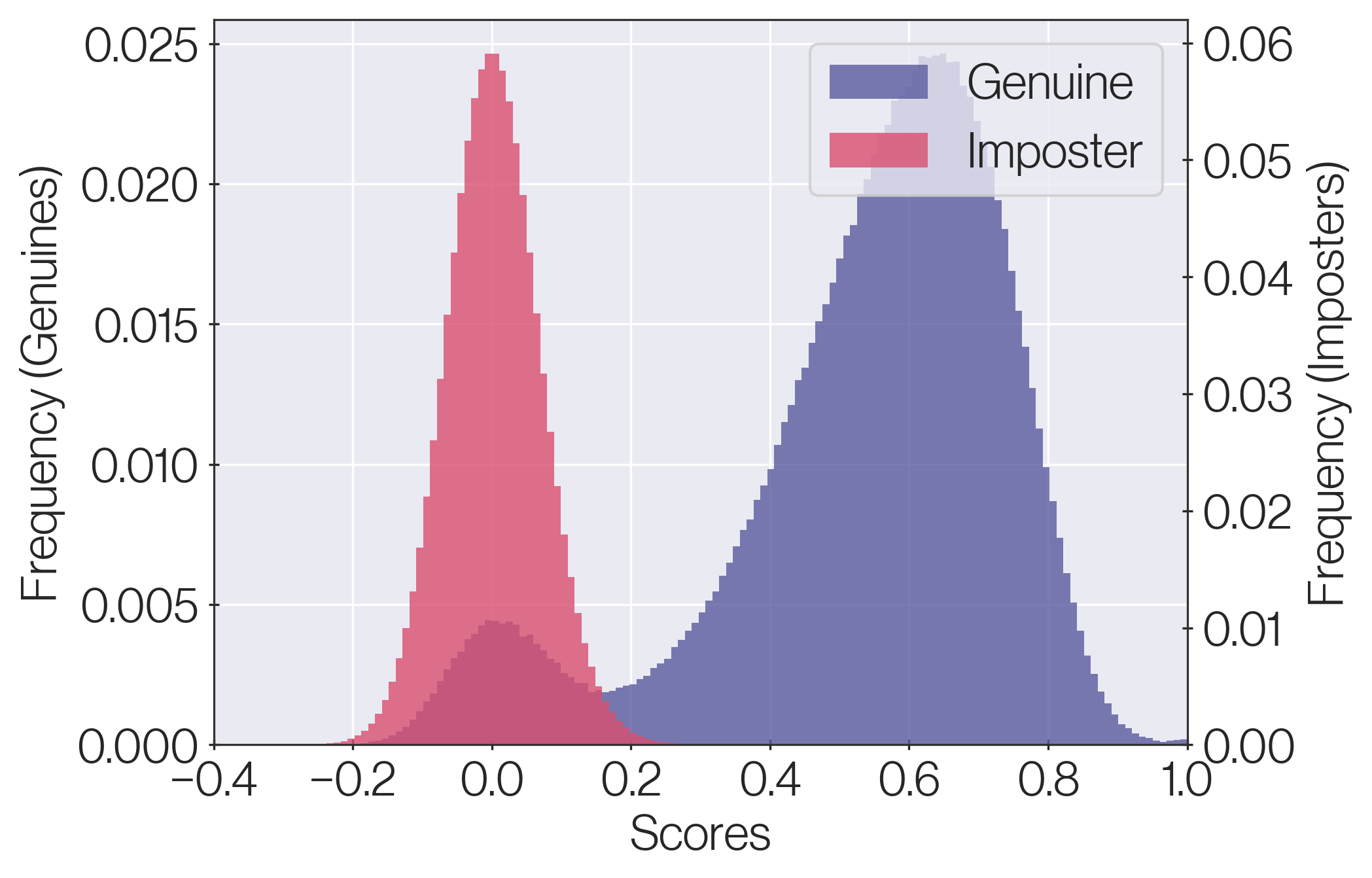} 
         \caption{CASIA-Webface}
         \label{fig:class_sep_casia_webface}
         
     \end{subfigure}
     \begin{subfigure}[t]{0.3\linewidth}
         \centering
         \includegraphics[width=\textwidth]{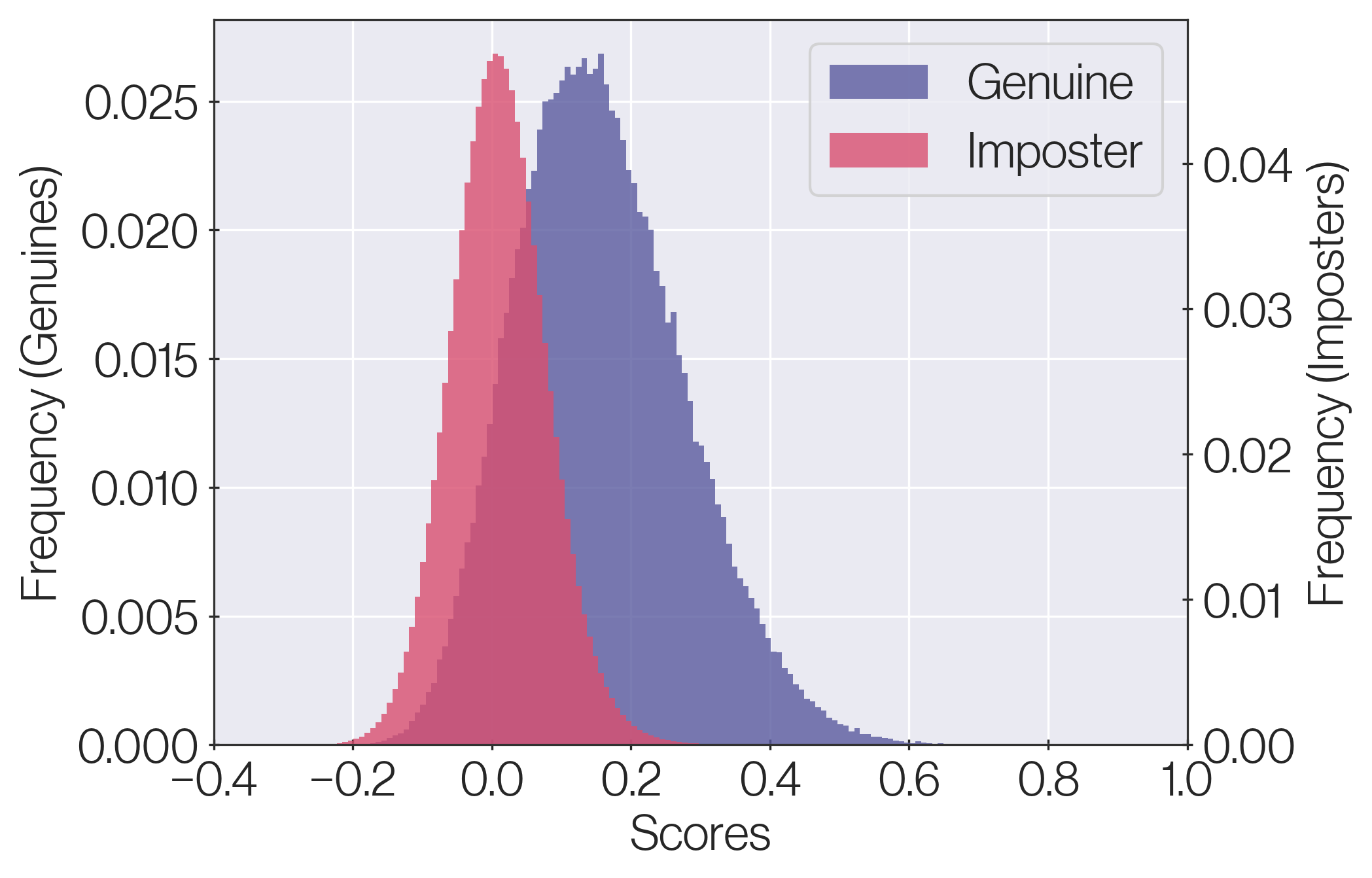} 
         \caption{SFace}
         \label{fig:class_sep_sg2cond}
     \end{subfigure}
     \begin{subfigure}[t]{0.3\linewidth}
         \centering
         \includegraphics[width=\textwidth]{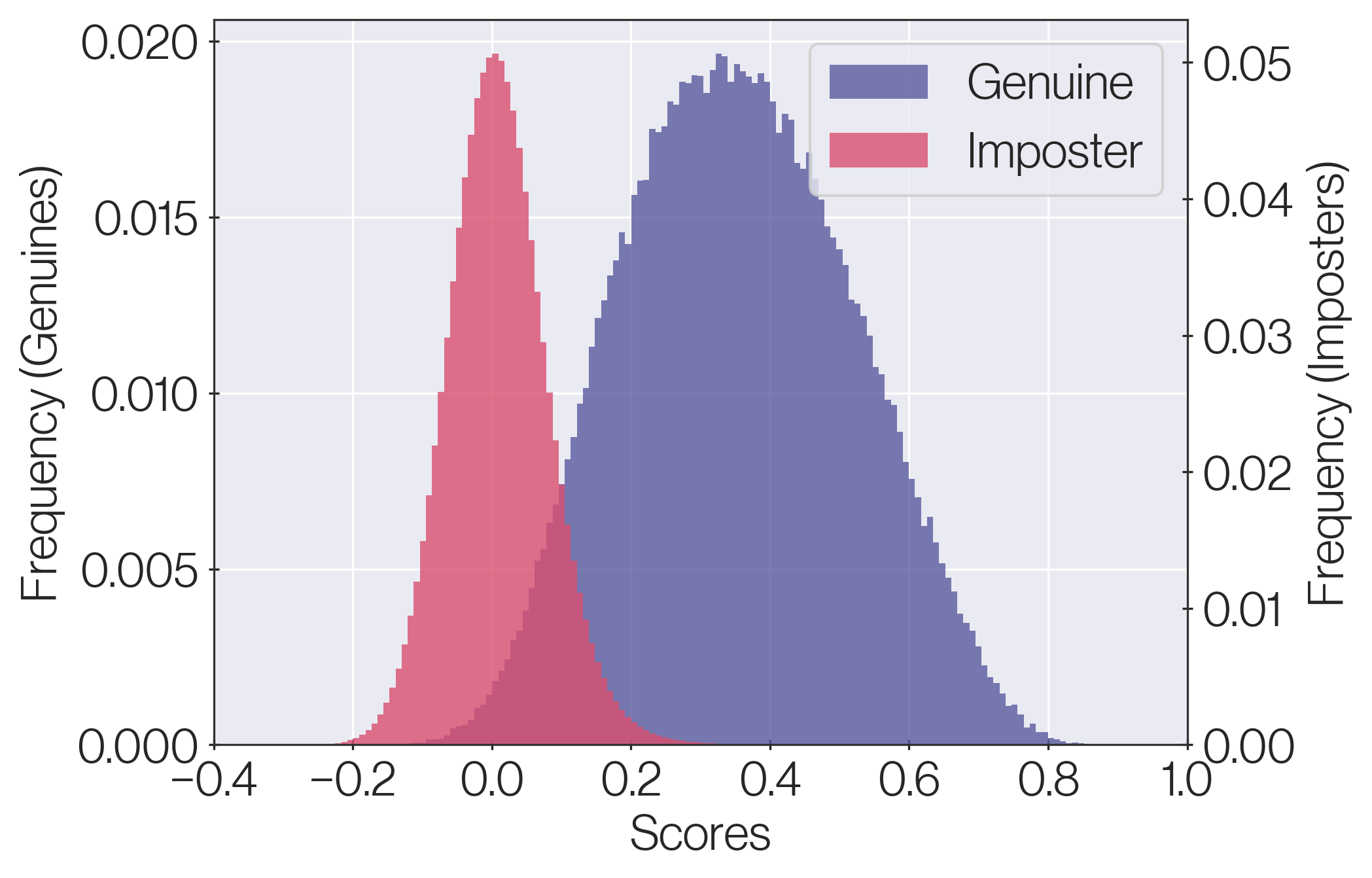}
         \caption{IDnet}
         \label{fig:class_sep_sg2idnet}
     \end{subfigure}
    \vspace{-3mm} 
      \caption{The genuine and imposter score distributions on respectively on the authentic dataset CASIA-Webface (a), on data created by SFace (b) and data created by IDnet (c). The small pump in the CASIA-Webface genuine distribution might be related to previously reported miss-labels \cite{DBLP:conf/iccv/WangWSWM19}. In comparison to SFace distributions, the distributions of IDnet are more similar to the authentic data in (a). }
    \label{fig:score_dist_separability}
    \vspace{-5mm}
    
\end{figure*}

\Cref{fig:score_dist_separability} shows the genuine and imposter score distributions of CASIA-Webface, SFace, and IDnet. In our solution IDnet (\cref{fig:class_sep_sg2idnet}) the genuine and imposter distribution are clearly more separated than our baseline SFace (\cref{fig:class_sep_sg2cond}) and are more similar to the distributions of the authentic data (CASIA-Webface, \cref{fig:class_sep_casia_webface}). This indicates that IDnet enforces the GAN to generate images with higher identity distinctiveness. Quantitatively this is shown in \Cref{tbl:similarity_stats} as EER, FMR100 and FMR1000. While the baseline SFace has a significantly higher EER, FMR100 and FMR1000 than CASIA-Webface, our solution IDnet results in more comparable values to CASIA-Webface. This shows that our solution generates data that possess higher identity discrimination than that of SFace.

\begin{table}
\centering
\footnotesize
\resizebox{\linewidth}{!}{%
\begin{tabular}{|l|c|c|c|} 
\hline
\multicolumn{1}{|c|}{}         & \textbf{EER} & \textbf{FMR 100} & \textbf{FMR 1000}  \\ 
\hline
CASIA-WebFace (ElasticFace)    & 0.062        & 0.084            & 0.118              \\
CASIA-WebFace (CurricularFace) & 0.063        & 0.085            & 0.120              \\ 
\hhline{|====|}
SFace (ElasticFace)            & 0.216        & 0.623            & 0.839              \\ 
\hline
SFace (CurricularFace)         & 0.226        & 0.628            & 0.848              \\ 
\hhline{|====|}
IDnet w/o DA (ElasticFace)     & 0.123        & 0.354            & 0.610              \\ 
\hline
IDnet w/o DA (CurricularFace)  & 0.128        & 0.354            & 0.604              \\ 
\hhline{|====|}
IDnet w/ DA (ElasticFace)      & 0.085        & 0.213            & 0.400              \\ 
\hline
IDnet w/ DA (CurricularFace)   & 0.085        & 0.204            & 0.380              \\
\hline
\end{tabular}
}
\vspace{-1mm}
\caption{Verification accuracy metrics indicating the class separability in each dataset, using two FR models. IDnet with DA is compared to IDnet without DA, to SFace \cite{boutros2022sface}, and to the authentic CASIA-WebFace. Note that IDnet produces the most similar results to the authentic data.}
\label{tbl:similarity_stats}
\vspace{-1mm}
\end{table}

\begin{table}
\centering
\resizebox{\linewidth}{!}{%
\begin{tabular}{l|c|c|c|c|c|c}
\multicolumn{1}{c|}{\textbf{Training Set}} & \textbf{LFW [\%]} & \textbf{AgeDB30 [\%]} & \textbf{CFP-FP [\%]} & \textbf{CA-LFW [\%]} & \textbf{CP-LFW [\%]} & \textbf{Avg. [\%]}  \\ 
\hline
IDnet-50 w/ DA                             & 84.83             & 63.58                 & 70.43                & 71.50                & 67.35                & 71.54               \\ 
\hline
IDnet-50 w/o DA                            & 81.33             & 58.48                 & 63.94                & 68.15                & 63.87                & 67.15              
\end{tabular}
}
\caption{Ablation on domain adaptation (DA) through Batch Normalization Statistics (BNS), as described in \Cref{sec:methodology}. FR models are trained on datasets of IDnet with and without DA with $50$ images per ID. The results show the benefit of the DA used in IDnet. Both sets contain 528K images (50 images/ID) and the training is performed without augmentation.}
\label{tbl:ablation_bns_results}
\vspace{-5mm}
\end{table}

\vspace{-4mm}
\paragraph{Intra-Class Variance:}

\begin{figure*}[ht!]
    \captionsetup[subfigure]{justification=centering}
     \centering
     \begin{subfigure}[t]{0.28\linewidth}
         \centering
         \includegraphics[width=\textwidth]{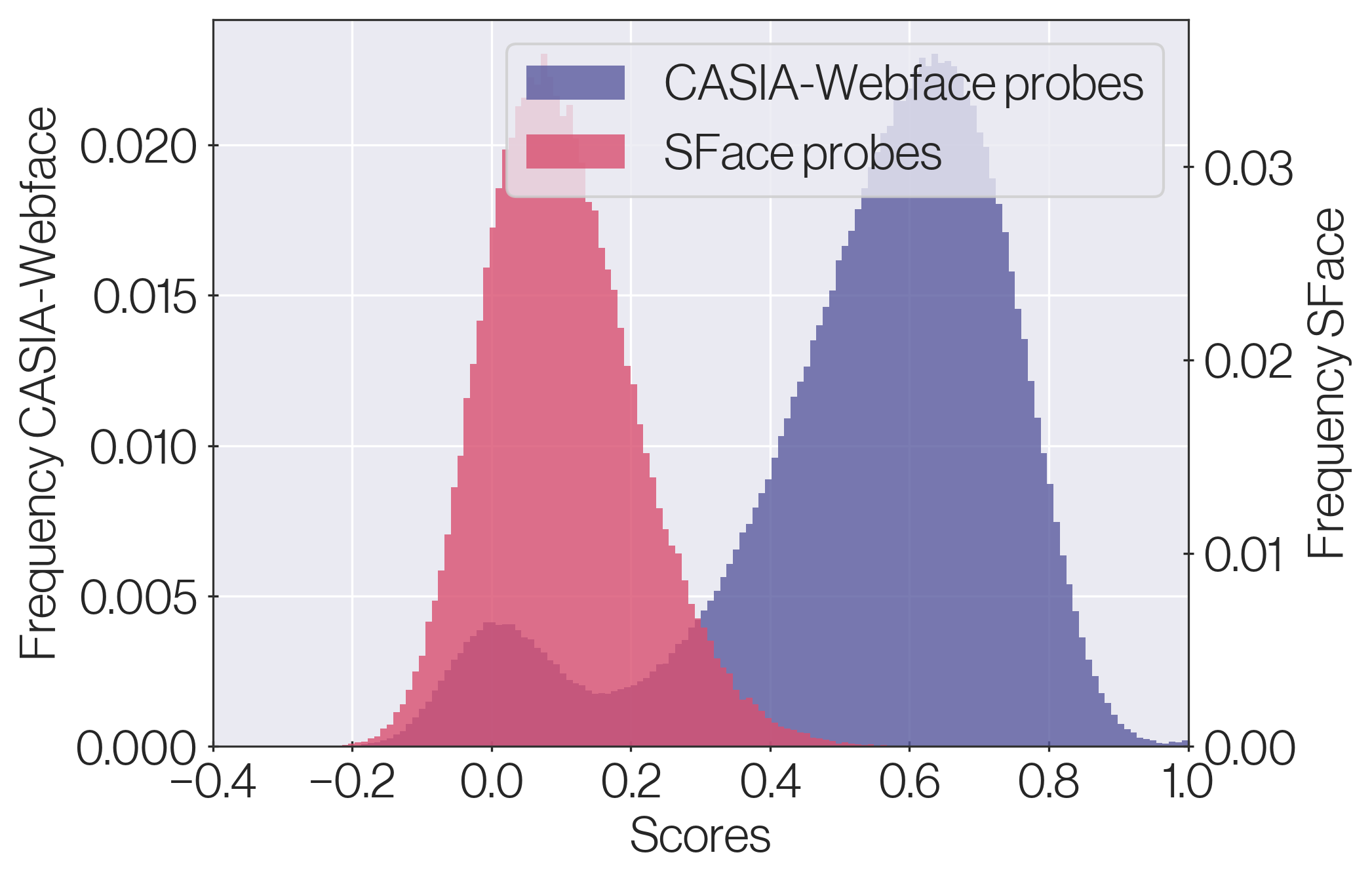}
         \caption{CASIA-Webface probes vs. SFace probe}
         \label{fig:id_sep_casia_condsg2}
         
     \end{subfigure}
     \begin{subfigure}[t]{0.28\linewidth}
         \centering
         \includegraphics[width=\textwidth]{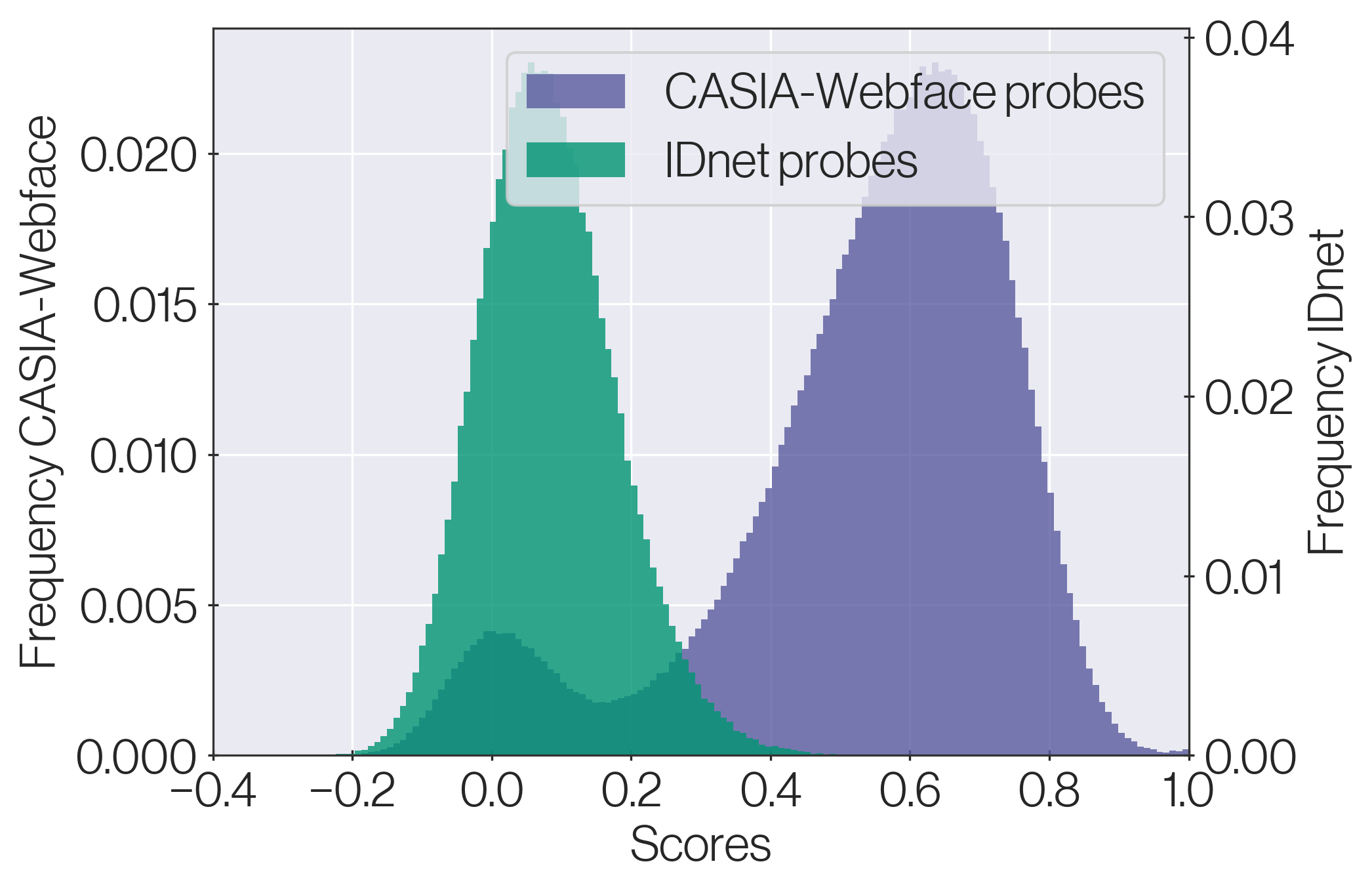}
         \caption{CASIA-Webface probes vs. IDnet probes.}
         \label{fig:id_sep_casia_sg2idnet}
     \end{subfigure}
     \begin{subfigure}[t]{0.28\linewidth}
         \centering
         \includegraphics[width=\textwidth]{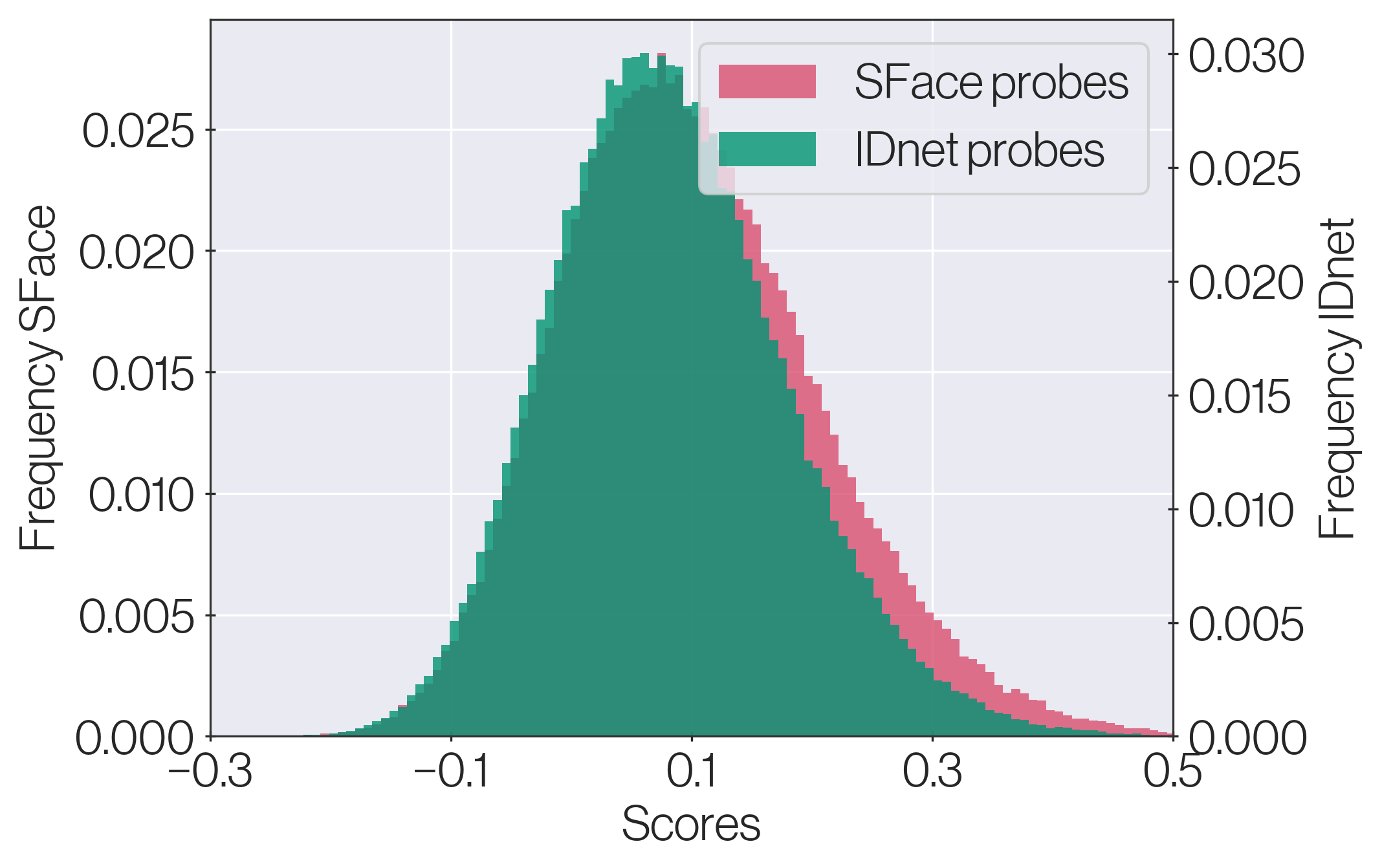}
         \caption{SFace probes vs. IDnet probes.}
         \label{fig:id_sep_condsg2_sg2idnet}
     \end{subfigure}
    \vspace{-3mm} 
      \caption{Score distributions between  CASIA-Webface references with probes with the same class label taken from CASIA-Webface (CASIA-Webface probes), SFace (SFace probes), and IDnet (IDnet probes). No significant overlap between distributions from authentic and synthetic probes is visible, indicating that SFace and IDnet are both generating synthetic identities with no significant overlap to authentic identities.}
    \label{fig:score_identity_similarity}
    \vspace{-4mm}
    
\end{figure*}

The $\text{ICV}$ score introduced in \Cref{sec:experimental_setup} is used to determine the variance between images of the same identity. CASIA-Webface scores $8.8 \times 10^{-4}$, our baseline SFace $14.9 \times 10^{-4}$ and our proposed solution IDnet $11.5 \times 10^{-4}$. While intra-class variation can be beneficial for the training of FR models, a very high ICV can also indicate that images of the same identity label are very different and that they may belong to more than one distinct identity. Therefore, the ICV value of face data used to train a FR system is a trade-off between intra- and inter-class variations, and thus a higher or a lower ICV value is not necessarily more beneficial. However, what matters here is to represent realistic conditions (i.e. similar to authentic data), which our IDnet achieves by scoring a more similar ICV to the authentic data when compared to that of the SFace baseline.
In \Cref{fig:id844_comparison} images of the same identity index, generated by SFace and our IDnet are shown. It is visually noticeable and quantitatively shown by the similarity scores that images of the identity generated by IDnet more distinctively maintain the targeted identity label.
This implies that the high $\text{ICV}$ score of SFace might be due to a lack of distinct identity information within images of the respective identity. The IDnet solution proposed in this work reduces this generative miss-labeling.
\vspace{-4mm}
\paragraph{Evaluating Similarity of Authentic and Synthetic Identities:}
\Cref{fig:score_identity_similarity} shows the genuine comparison score distributions resulting from comparison pairs that consist of CASIA-Webface references and probes of the same class label either from authentic CASIA-Webface (blue), synthetic SFace (red), or IDnet (green).
If the comparison scores of involving synthetic probes (red or green) lay in the same range of that of the authentic probes (blue), then one would conclude that the comparison scores between the authentic and synthetic images of the same labels are high enough to be considered genuine comparison, thus there is an identity link between the authentic and synthetic data.
As \Cref{fig:id_sep_casia_condsg2} shows, no significant overlap is present when comparing probes of CASIA-Webface with probes synthesized with SFace. The same can be observed for probes generated with IDnet (\cref{fig:id_sep_casia_sg2idnet}). 
This implies that neither for SFace nor for IDnet the identity information of the authentic identity is strongly present in the synthetic identity, confirming the outcome of \cite{boutros2022sface}. 
From this, it is concluded that adding the IDnet in the generation process also ensures that synthetic identities, not linked to the authentic ones, are generated. Therefore, the solution complies with the privacy-driven motivations behind our work.

\Cref{fig:id_sep_condsg2_sg2idnet} shows the comparison score distributions when the probes are from SFace and IDnet. The distribution of IDnet is slightly shifted to the lower similarity range in comparison to the SFace distribution. This indicates that there is a slightly smaller identity link between  IDnet faces (in comparison to the ones from SFace) and their respective authentic identities. 
Our proposed solution, therefore, provides an additional improvement for generating purely synthetic identities over the baseline SFace.

\vspace{-4mm}
\paragraph{The benefit of DA:} To prove the validity of including the BNS DA in the IDnet design, we trained two instances of the IDnet, with and without the DA. We first show the effect on the genuine-imposter separability of the generated data in Table \ref{tbl:similarity_stats}. The tables shows that the IDnet data without DA do produce higher EER, FMR100, and FMR1000 in comparison to the IDnet with DA. Additionally, as shown in Table \ref{tbl:ablation_bns_results}, the FR models trained on the IDnet data generated with the DA achieved higher face verification performance across all considered benchmarks, in comparison to the one trained on data generated by the IDnet without DA. This proves the validity of our choice to include the DA in the IDnet design.

\begin{table}[ht!]
\centering
\resizebox{\linewidth}{!}{%
\begin{tabular}{c|c|c|c|c|c|c|c|c}
\textbf{Training Set}    & \textbf{Images} & \textbf{\#Images / ID} & \textbf{LFW [\%]}  & \textbf{AgeDB30 [\%]} & \textbf{CFP-FP [\%]} & \textbf{CA-LFW [\%]} & \textbf{CP-LFW [\%]} & \textbf{Avg. [\%]}  \\ 
\hline
CASIA-WebFace            & 494K            & 46                     & 99.55              & 94.55                 & 95.31                & 93.78                & 89.95                & 94.63               \\ 
\hline
SFace-10  \cite{boutros2022sface}               & 105K            & 10                     & 87.13              & 63.30                 & 68.84                & 73.47                & 66.82                & 71.91               \\
SFace-20  \cite{boutros2022sface}                   & 211K            & 20                     & 90.50              & 69.17                 & 73.33                & 76.35                & 71.17                & 76.10               \\
SFace-40  \cite{boutros2022sface}                   & 423K            & 40                     & 91.43              & 69.87                 & 73.10                & 76.92                & 73.42                & 76.95               \\
SFace-60  \cite{boutros2022sface}                   & 634K            & 60                     & 91.87              & 71.68                 & 73.86                & 77.93                & 73.20                & 77.71               \\ 
\hline
USynthFace \cite{usynthface}              & 100K            & 1                      & 91.52              & 69.30                 & 78.46 (3)            & 75.35                & 71.93                & 77.31               \\
USynthFace \cite{usynthface}                & 200K            & 1                      & 91.93              & 71.23                 & 78.03                & 76.73                & 72.27                & 78.04               \\
USynthFace \cite{usynthface}                & 400K            & 1                      & 92.23              & 71.62                 & 78.56 (2)            & 77.05                & 72.03                & 78.30               \\ 
\hline
SynFace \cite{synface}                 & 500K            & 50                     & 91.93              & 61.63                 & 75.03                & 74.73                & 70.43                & 74.75               \\ 
\hline
DigiFace-1M \cite{bae2023digiface}             & 500K            & 50                     & 95.40 (1) &       76.97 (1)    &          87.40 (1)               & 78.62                & 78.87 (1)   &        83.45 (1)  \\
\hline
DigiFace-1M (No Aug.)  \cite{bae2023digiface}    & 500K            & 50                     & 88.07              & 60.92                 & 70.99                & 69.23                & 66.73                & 71.19               \\ 
IDnet-50 (No Aug.) [Our] & 528K            & 50                     & 84.83              & 63.58                 & 70.43                & 71.50                & 67.35                & 71.54               \\
\hline
IDnet-10 [Our]           & 105K            & 10                     & 92.68 (3)          & 74.42 (3)             & 74.73                & 81.92 (1)   & 74.32                & 79.61 (3)           \\
IDnet-20 [Our]           & 211K            & 20                     & 92.58              & 74.78 (2)             & 76.34                & 80.72 (2)            & 75.77 (2)            & 80.04 (2)           \\
IDnet-40 [Our]           & 423K            & 40                     & 92.88 (2)          & 73.37                 & 76.90                & 79.42                & 74.98 (3)            & 79.51               \\
IDnet-50 [Our]           & 528K            & 50                     & 92.58              & 73.53                 & 75.40                & 79.90 (3)            & 74.25                & 79.13               \\
IDnet-60 [Our]           & 634K            & 60                     & 92.30              & 73.67                 & 75.93                & 79.40                & 74.35                & 79.13   \\ \hline           
\end{tabular}
}
\vspace{-2mm}
\caption{FR accuracy when trained on IDnet  with $10$ (IDnet-10), $20$ (IDnet-20), $40$ (IDnet-40), and $50$ (IDnet-50) synthetic images per identity on mainstream benchmarks, respectively, compared to other work. The top three performing solutions are indicates as (1), (2), and (3). Results obtained without data augmentation are labeled as (No Aug.).}
\label{tbl:verifiation_results}
\vspace{-4mm}
\end{table}

\vspace{-4mm}
\paragraph{Verification Performance of FR trained on IDnet data:}
The target application of our synthetic face images is the training of FR models. The results on the mainstream benchmark for the training of a ResNet-50 backbone (as detailed in \cref{sec:experimental_setup}) are shown in \Cref{tbl:verifiation_results}. 
Investigating involving a different number of samples per identity in the FR training, we notice that increasing the number of images per identity to more than 10 does not drastically affect the FR performance. In comparison, the SFace showed a clear increase in performance when the number of images per identity is increased beyond 10. This might be due to the higher identity discriminant nature of our IDnet (in comparison to SFace) as discussed in \Cref{sec:results} which means that fewer images are required to represent a correct class center for each training identity.

When compared to the baseline SFace \cite{boutros2022sface}, our proposed IDnet achieves a significantly higher accuracy on all mainstream benchmarks, specifically for scenarios with fewer images per identity. 
Likewise, IDnet outperforms other related works like USynthFace \cite{usynthface} and SynFace \cite{synface}. 
This emphasizes that the IDnet component allows the generation of face images that retain the selected identity information in a manner that better mimics that of the authentic data.\\
While DigiFace-1M \cite{bae2023digiface} achieves higher accuracies on nearly all benchmarks, IDnet achieves the best performance on the CA-LFW benchmark and comes as a very close second in the AgeDB30 (both targeting age-gap evaluations).
This might be linked to a major factor that boosts the results in DigiFace-1M \cite{bae2023digiface}, i.e. the aggressive augmentation. 
While our solution insures natural variations (including age), the augmentations introduced in \cite{bae2023digiface} to boost the performance do not affect the cross-age performance as other face variations.
Given that the detailed augmentation parameters are not specified (or released publicly) in \cite{bae2023digiface}, a fairer comparison would be between FR models trained without augmentation on our IDnet and the DigiFace-1M. In this comparison (in \Cref{tbl:verifiation_results}), our IDnet outperforms DigiFace-1M in the average accuracy and very significantly on the cross-age benchmarks, CA-LFW and AgeDB30.
Additionally, the performance of DigiFace-1M comes with great limitations. The images of DigiFace-1M are rendered with the physically-based-rendering engine Cycles \cite{bae2023digiface}, utilizing $300$ NVIDIA M60 GPU for $10$ days. In comparison, $500k$ images are generated by IDnet on a single Nvidia GeForce RTX 6000 GPU in less than 2 hours. 
Combining our achieved results without and with the open source augmentations along with the significantly faster synthesis the strength of our solution compared to previous work, including DigiFace-1M, becomes evident.

\section{Conclusion}

In this work, we presented a novel identity conditional three-player GAN, IDnet, which enables synthetic image  generation of synthetic identities with high  variability and strong identity separability. We empirically demonstrated that utilizing our IDnet to train FR model achieved relatively high verification accuracies on the main FR benchmarks,  outperformed previous GAN-based approaches, and achieved competitive results to the computationally costly digital rendering-based synthetic image approach.  
As concluding remarks, this work accelerates the switch towards training FR models in a privacy-aware manner and explores a new research direction for incorporating identity information in the generation process.

\textbf{Acknowledgment}
This research work has been funded by the German Federal Ministry of Education and Research and the Hessian Ministry of Higher Education, Research, Science and the Arts within their joint support of the National Research Center for Applied Cybersecurity ATHENE. This work has been partially funded by the German Federal Ministry of Education and Research (BMBF) through the Software Campus Project.

{\small
\bibliographystyle{ieee_fullname}

}

\end{document}